\documentclass[letterpaper, 10 pt, conference]{ieeeconf}
\IEEEoverridecommandlockouts
\usepackage{cite}
\usepackage{amsmath,amssymb,amsfonts}
\usepackage{algorithmic}
\usepackage{graphicx}
\usepackage{textcomp}
\usepackage{xcolor}
\def\BibTeX{{\rm B\kern-.05em{\sc i\kern-.025em b}\kern-.08em
    T\kern-.1667em\lower.7ex\hbox{E}\kern-.125emX}}

\usepackage{siunitx}
\usepackage{url}
\usepackage{overpic}
\usepackage{pgf}
\usepackage{microtype}
\usepackage{booktabs}
\usepackage{color, colortbl}
\usepackage{tikz}
\usepackage{tensor}

\newcommand{\X}{\mathcal{X}}
\newcommand{\Z}{\mathcal{Z}}

\newcommand{\Figure}{Fig.~}

\newcommand{\Equation}{Eq.~}

\newcommand{\ie}{{i.e.,~}}

\newcommand{\SO}{\mathrm{SO}}
\newcommand{\SOthree}{\SO(3)}
\newcommand{\Real}{\mathbb{R}}

\newcommand{\calC}{{\cal C}}

\newcommand{\calI}{{\cal I}}

\newcommand{\calL}{{\cal L}}

\newcommand{\calP}{{\cal P}}

\newcommand{\calT}{{\cal T}}

\newcommand{\calX}{{\cal X}}
\newcommand{\calZ}{{\cal Z}}
\newcommand{\sat}{\mathbf{s}}
\newcommand{\ecef}{\mathtt{E}}

\newcommand{\R}{\mathbf{R}}

\newcommand{\T}{\mathbf{T}}

\newcommand{\residual}{\mathbf{r}}

\newcommand{\tran}{\mathbf{p}}
\newcommand{\vel}{\mathbf{v}}
\newcommand{\bias}{\mathbf{b}}

\newcommand{\State}{\boldsymbol{x}}

\newcommand{\Base}{\mathtt{{B}}}

\newcommand{\Lidar}{\mathtt{L}}
\newcommand{\World}{\mathtt{W}}
\newcommand{\Imu}{\mathtt{I}}

\newcommand{\Anchor}{\mathtt{{A}}} %
\newcommand{\lightspeed}{c}

\newcommand{\base}{\mathtt{{B}}}

\newcommand{\world}{\mathtt{{W}}}
\newcommand{\imu}{\mathtt{{I}}}

\DeclareMathOperator*{\argmax}{arg\,max}
\DeclareMathOperator*{\argmin}{arg\,min}

\newcommand{\SEthree}{\ensuremath{\mathrm{SE}(3)}}

\makeatletter
\def\underbracex#1#2{\mathop{\vtop{\m@th\ialign{##\crcr
   $\hfil\displaystyle{#2}\hfil$\crcr
   \noalign{\kern3\p@\nointerlineskip}%
   #1\crcr\noalign{\kern3\p@}}}}\limits}

\def\upbracefilla{$\m@th \setbox\z@\hbox{$\braceld$}%
  \bracelu\leaders\vrule \@height\ht\z@ \@depth\z@\hfill
\kern\p@\vrule \@width\p@\kern\p@\vrule \@width\p@\kern\p@\vrule \@width\p@
$}

\def\upbracefillb{$\m@th \setbox\z@\hbox{$\braceld$}%
\vrule \@width\p@\kern\p@\vrule \@width\p@\kern\p@\vrule \@width\p@\kern\p@
 \leaders\vrule \@height\ht\z@ \@depth\z@\hfill\bracerd
  \braceld\leaders\vrule \@height\ht\z@ \@depth\z@\hfill
\kern\p@\vrule \@width\p@\kern\p@\vrule \@width\p@\kern\p@\vrule \@width\p@
$}

\def\upbracefilld{$\m@th \setbox\z@\hbox{$\braceld$}%
\vrule \@width\p@\kern\p@\vrule \@width\p@\kern\p@\vrule \@width\p@\kern\p@
 \leaders\vrule \@height\ht\z@ \@depth\z@\hfill\bracerd
  \braceld\leaders\vrule \@height\ht\z@ \@depth\z@\hfill
 \braceru$}

\makeatother

\title{\LARGE \bf Factor Graph Fusion of Raw GNSS Sensing with IMU and
Lidar \\ for Precise Robot Localization without a Base Station}

\author{Jonas Beuchert$^{1,2}$, Marco Camurri$^{2,3}$, and Maurice Fallon$^{2}$%
\thanks{Funding: EPSRC Centre for Doctoral Training in Autonomous Intelligent Machines \& Systems (J. Beuchert), Royal Society Univ. Research Fellowship (M. Fallon), EU Horizon Europe Project DigiForest (M. Camurri). For the purpose of open access, the authors have applied a Creative Commons Attribution (CC BY) license to any Accepted Manuscript version arising.}
\thanks{$^{1}$Dept. of Computer
Science, Univ. of Oxford, UK}
\thanks{$^{2}$Oxford Robotics Inst., Dept. of Eng.
Science, Univ. of Oxford, UK}
\thanks{$^{3}$Faculty of Science \& Technology, Free Univ. of Bozen-Bolzano, Italy}
\thanks{ \tt\footnotesize \{beuchert,mfallon\}@robots.ox.ac.uk, marco.camurri@unibz.it}
}

\begin{document}
\maketitle
\thispagestyle{empty}
\pagestyle{empty}

\begin{abstract}
Accurate localization is a core component of a robot's navigation system. To
this end, global navigation satellite systems (GNSS) can provide absolute
measurements outdoors and, therefore, eliminate long-term drift. However, fusing
GNSS data with other sensor data is not trivial, especially when a robot moves
between areas with and without sky view. We propose a robust approach that
tightly fuses raw GNSS receiver data with inertial measurements and, optionally,
lidar observations for precise and smooth mobile robot localization. A factor
graph with two types of GNSS factors is proposed. First, factors based on
pseudoranges, which allow for global localization on Earth. Second, factors
based on carrier phases, which enable highly accurate relative localization,
which is useful when other sensing modalities are challenged. Unlike traditional
differential GNSS, this approach does not require a connection to a base
station. 
On a public urban driving dataset, our approach achieves accuracy
comparable to a state-of-the-art algorithm that fuses visual inertial odometry
with GNSS data---despite our approach not using the camera, just inertial and
GNSS data. We also demonstrate the robustness of our approach using data from a
car and a quadruped robot moving in environments with little sky visibility,
such as a forest. The accuracy in the global Earth frame is still
\SIrange{1}{2}{\meter}, while the estimated trajectories are discontinuity-free
and smooth. We also show how lidar measurements can be tightly integrated. We
believe this is the first system that fuses raw GNSS observations (as opposed to
fixes) with lidar in a factor graph.
\end{abstract}

\section{INTRODUCTION}
A key enabler for autonomous navigation is accurate localization using only a
robot's onboard sensors. Effective sensor fusion is essential to maximize
information gain from each sensing modality. For example, proprioception based
on inertial measurement units (IMUs) or encoders together with exteroception
from cameras or lidars can be used to achieve a smooth, but slowly drifting,
estimate of the motion of a mobile platform in a local environment. In contrast,
satellite navigation can be used to estimate positions in a global Earth frame.
These estimates are drift-free, but require sky visibility. Therefore, fusion of
satellite navigation, proprioception, and exteroception is desirable for
long-term autonomy.

\begin{figure}
\centering
\includegraphics[width=0.308\columnwidth]{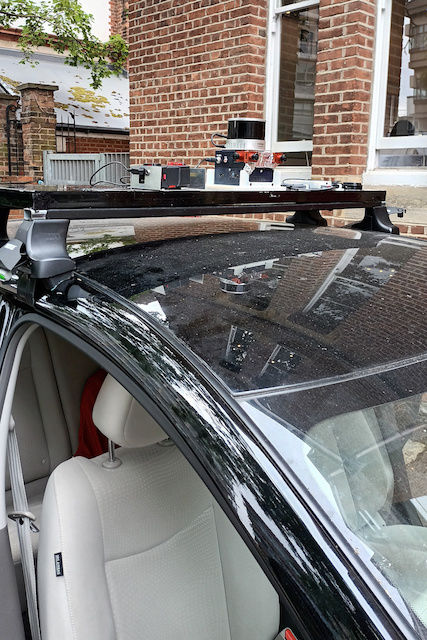}
\includegraphics[width=0.308\columnwidth]{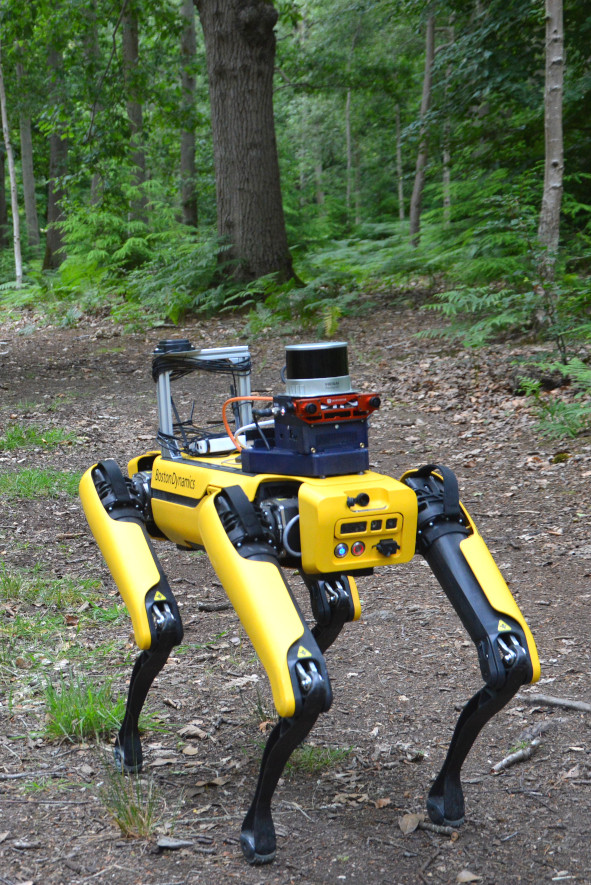}
\includegraphics[width=0.308\columnwidth]{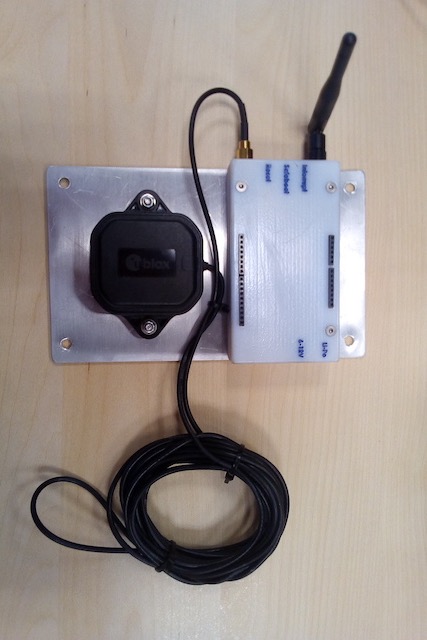}

\caption{Setups for experimental validation. \textbf{Left:} a car driving on
public roads. \textbf{Center:} a quadruped robot moving in urban and forest
environments. \textbf{Right:} a GNSS receiver carried handheld. Raw GNSS
measurements were tightly fused with inertial sensing and optionally lidar using
factor graphs.}
\label{fig:fg_art}
\end{figure}

The most popular way to fuse data from global navigation satellite systems
(GNSS) with onboard perception sensing is two-stage fusion. First, an
independent estimator running on the GNSS receiver computes global position
fixes from raw GNSS observations. These fixes are then used as position priors
within a second-stage pose estimator with proprioceptive and exteroceptive
measurements~\cite{shan2020lio, gong2020graph}. However, this two-stage fusion
has several disadvantages, e.g., the two stages are only loosely coupled, which usually causes lower accuracy and higher
uncertainty, especially, when a receiver acquires only a few satellites, or if the internal estimator dynamics of the first stage are unknown.
Furthermore, maintaining a persistent network connection for differential GNSS
(DGNSS) is impossible or inconvenient in some applications. Without DGNSS,
receiver position uncertainty is in the order of several meters.
Because of this, GNSS can only be used to anchor a robot's trajectory in the global
Earth frame or to correct long-term drift, but cannot aid accurate local
navigation, e.g., if the robot's exteroception fails.

We follow an alternative approach that addresses these disadvantages by instead
fusing the raw observations of the individual satellites with proprioceptive and
exteroceptive measurements in a single state estimator, which is more tightly
coupled. This allows us to leverage information from even just a few satellites
(e.g., less than four)---information which would be discarded in the two-stage
approach.

For each visible satellite and a given frequency band, a conventional GNSS
receiver can provide three types of observables: pseudoranges, carrier phases,
and Doppler shifts. These measurements are affected by a number of errors,
rendering fusion particularly challenging and requiring robust optimization. In
addition, the different observables have distinct properties: pseudoranges have
a comparably high uncertainty while carrier phases are accurate, but challenging
to use in real time without a permanent communication link to a base station.
Our algorithm addresses both challenges.

Specifically, the contributions of our work are:
\begin{itemize}
\item A novel factor graph design incorporating two types of raw GNSS
observables (pseudroanges for drift-free localization in the global Earth frame
and differential carrier phases for accurate and smooth local positioning)
together with inertial measurements and optionally lidar---to our knowledge the
first factor graph design to do so.
\item A single real-time optimization phase, which implicitly handles GNSS
initialization, normal operation, and GNSS drop-out. This eliminates the need to
switch between different modes for the aforementioned phases and leads to fast
convergence.
\item An extensive evaluation on a car and a quadruped robot moving in
challenging scenarios and comparison against the state-of-the-art on a public
dataset.
\end{itemize}

\section{RELATED WORK}
Fusion of individual GNSS satellite observations (rather than
pre-computed GNSS fixes) with proprioceptive and exteroceptive measurements in a
single estimation framework has been pursued in previous work.
Traditional methods for integration of GNSS and proprioception, e.g, inertial or encoder measurements, often used filter-based estimation~\cite{farrell2008, groves2013}.
In contrast, some recent approaches from the robotics community leveraged factor graph optimization.
Factor graphs are a popular estimation framework for various fusion problems in robotics~\cite{Dellaert2017} and it would be desirable to be able to incorporate raw GNSS observations in this manner.
Most approaches used the pseudoranges provided
by the GNSS receiver~\cite{watson2018evaluation, watson2019diss, wen2019tightly,
wen2020time, sunderhauf2012towards, sunderhauf2013switchable, gong2019tightly}.
A pseudorange is the observed travel time of a radio signal from a satellite to
the receiver multiplied by the speed of light. They can be seen as range-only
observations of distant landmarks; although, pseudoranges usually have much
larger meter-scale uncertainties than the centimeter-scale ones of visual
landmarks~\cite{gong2019tightly}. For example, Gong et al., Wen et al., and Cao
et al. fused pseudoranges with combinations of visual or inertial measurements
to show that a tightly coupled approach can be robust in urban canyon scenarios
with limited sky visibility~\cite{gong2019tightly, wen2019tightly, cao2022tro}.
They achieved mean positioning errors of a few meters. Due to their meter-scale
measurement uncertainty, pseudoranges cannot be employed for centimeter-accurate
localization---only for anchoring a trajectory in a global Earth frame and
eliminating odometry drift. For accurate local navigation, these approaches rely
on a combination of proprioception and exteroception.

To overcome this limitation, carrier-phase observations can also be used. The
satellites transmit their data on carrier signals in the gigahertz range, i.e.,
sine waves with fixed frequencies. If the receiver could measure the number of
sine-wave periods between the satellite and itself, this would serve as an
additional observation that is proportional to the receiver-satellite distance.
It would be more accurate than the pseudorange because signal wavelengths are in
the range \SIrange{18}{26}{\centi\metre}. However, the receiver cannot count the
absolute number of sine-wave periods. Instead, it can observe the phase of the
carrier wave and count the change in the number of waves since the receiver
first locked onto the signal. (The sum of these values is usually referred to as
the \textit{carrier phase}.) Thus, the range observation can only be inferred up
to an unknown integer number of wavelengths, known as the \textit{integer
ambiguity}, which is the number of full waves when the signal was locked.
Usually, real-time kinematic positioning (RTK) is used to resolve the integer
ambiguity in real time. However, this requires a persistent connection between
the moving GNSS receiver (the \textit{rover}) and a nearby stationary second
receiver at a known location (the \textit{base}).

In the literature, approaches have been described that could make use of carrier
phases in real time \textit{without the need for a base station}. For example,
Suzuki used carrier phases to create factors between states at different times
to obtain relative distance measurements w.r.t. past
epochs~\cite{suzuki2020time}. The author named this method \textit{time-relative
RTK} because of its similarity to RTK---with the current observations as rover
observations and a set of previous observations as observations to a virtual
base station. For each continuously tracked satellite, the approach estimated
the integer ambiguity using the LAMBDA method~\cite{teunissen1995lambda,
takasu2009development}. If the method could not resolve the integer ambiguity,
then no factor was added to the graph. In this way, the integer ambiguity
estimation was not tightly coupled with the factor-graph-based state estimation.
Combining these carrier-phase factors with pseudorange and Doppler-shift factors
in a single factor graph, Suzuki achieved mean positioning errors of
\SIrange{2}{5}{\centi\meter} after UAV flights over \SI{200}{\meter} or
\SI{100}{\second} with good sky visibility and post-processing of data from a
single-frequency receiver. There was no real-time evaluation of this method and
it did not address fusion with non-GNSS measurements.

Lee et al. described a second method called \textit{sequential-differential
GNSS} in which they also create differential carrier-phase factors between
different states in time~\cite{Lee2022ICRA}. This approach also canceled the
integer ambiguities. It fused the carrier phases with pseudorange, Doppler
shift, visual, and inertial measurements in a multi-state constraint Kalman
filter (MSCKF). However, they also used time-relative factors for the
pseudoranges. They anchored the local trajectory in the global Earth frame
during an initialization phase and afterwards only used the pseudorange and
carrier-phase observations for relative localization w.r.t. previously estimated
states. This limits the usefulness of pseudorange observations for reducing
\textit{long-term} drift, especially, if sky visibility is lost intermittently
or very limited. They demonstrated an RMSE of \SI{0.32}{\meter} on a handheld
dataset where sky visibility was never interrupted and differential GNSS factors
could always be created between the current and previous state.

In summary, no method has been presented yet that tightly fuses pseudoranges and
carrier phases with proprioception and/or exteroception for \textit{long-term}
autonomous localization \textit{in real time} using a single GNSS receiver.
Furthermore, to the best of our knowledge, tight factor graph fusion of raw GNSS data with
lidar has not been addressed in the literature.

\section{PROBLEM STATEMENT}
We aim to estimate the pose of a mobile platform that is equipped with a
GNSS receiver, an IMU, and optionally a lidar---in real time and in a global
Earth frame. The estimated trajectory is required to be smooth/discontinuity-free.

\subsection{Frames}
\label{sec:frames}

\begin{figure}
\centering
\includegraphics[width=\columnwidth]{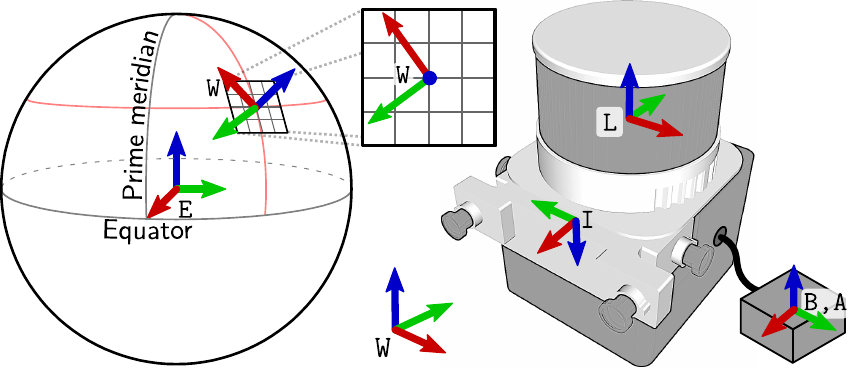}
\caption{Reference frames: the Earth-centered-Earth-fixed
(ECEF) frame $\ecef$, the local
frame $\World$, and
the frames on the platforms, including
base $\Base$, which coincides with the GNSS antenna frame
$\Anchor$, IMU frame $\Imu$,
and lidar frame $\Lidar$. %
}
\label{fig:frames}
\end{figure}

Ultimately, we are interested in position estimates in geographic coordinates
(latitude, longitude). However, for computational reasons, we internally use the
Cartesian Earth-centered-Earth-fixed (ECEF) frame $\ecef$ as global frame. Its
$z$-axis is the Earth's rotation axis, and its $x$-axis points to the prime
meridian, cf. Fig.~\ref{fig:frames}.

If the system has exteroceptive sensors, then we are not only interested in a
global position estimate, but also in maintaining a smooth local trajectory with
no discontinuities during initial GNSS convergence. Therefore, we establish a
local frame~$\World$ that we align with the platform's pose at the start of the
experiment.
We estimate the transformation~$\T_{\ecef \World} \in \SOthree \times \Real^3$ between local frame~$\World$ and global frame~$\ecef$ jointly with the platform position in frame~$\World$.
This ensures that the position estimate in~$\World$ remains smooth even in scenarios where~$\T_{\ecef \World}$ converges slowly due to no or little GNSS observations being available at the start of an experiment.

If the sensing system has only GNSS and inertial sensing, then we set $\World$
to be identical to~$\ecef$. The frames rigidly attached to the robot are base
$\Base$ (coincident with the GNSS antenna frame $\Anchor$), IMU $\Imu$, and
lidar $\Lidar$, see Fig.~\ref{fig:frames}.

\subsection{State}
The state of the system at time $t_i$ is
\begin{equation}
\State_i \triangleq \left[\R_i,\tran_i,\vel_i, \bias^{\mathrm{g}}_i \,\;
\bias^{\mathrm{a}}_i, \delta\mathbf{t}^\mathrm{r}_i \right] \in \SOthree \times
\Real^{12 + M},
\end{equation}
where $\R_i = \R_{\world\base}(t_i) \in \SOthree$ and $\tran_i = \tensor[_{\world}]{\tran}{_{\world\base}}(t_i) \in \Real^3$ are the orientation and position of the base $\base$ with respect to $\world$, respectively; $\vel_i = \tensor[_\base]{\vel}{_{\world\base}}(t_i) \in \Real^3$ is the linear velocity of $\base$ with respect to $\World$ expressed in $\base$. The IMU's slowly changing gyroscope and accelerometer biases 
$\bias^{\mathrm{g}}_i, \bias^{\mathrm{a}}_i \in \Real^3$ are expressed in frame $\imu$.
Finally, $\delta\mathbf{t}^\mathrm{r}_i \in \Real^M$ is the vector of the clock
offsets between each satellite system and the receiver clock. Our receiver
accesses four satellite systems (\ie $M = 4$): GPS, BeiDou, GLONASS, and
Galileo.

If the setup includes a lidar, we estimate the transformation~$\T_{\ecef
\World}$ in addition to the system states~$\State_i$. The history of all
unknowns is then
\begin{equation}
    \X_k \triangleq \left\{\left\{\State_i \right\}_{ i \in
\calT_k}, \T_{\ecef \World} \right\},
\end{equation}
where $\calT_k$ is the set of all state indices in a fixed-length time window up to time~$t_k$.

\subsection{Measurements}
The measurements include the proper acceleration and angular velocity in the IMU
frame~$\Imu$ and lidar point clouds $\calL_{ij}$. Inertial measurements are
received as a set between times $t_i$ and $t_j$ as $\calI_{ij}$. They are preintegrated after gravity/bias compensation, as explained in Sec.~\ref{sec:imu-factors}. The GNSS
receiver observes pseudoranges $\calP_i$ and double-differenced carrier phases
$\calC_{ij}$. (The latter are explained in Sec.~\ref{sec:cp}.) Thus, all
measurements in a time window $\calT_k$ are
\begin{equation}
    \Z_k \triangleq \left\{ \calI_{ij}, \calL_{ij}, \calP_i, \calC_{ij} \right\}_{
i,j \in \calT_k}.
\end{equation}
We create a state $\State_i$ whenever there is a GNSS observation at time $t_i$
and motion correct a potentially received lidar point cloud at that
timestamp~\cite{wisth2022tro}. If there is no GNSS observation for a certain
amount of time, e.g., due to no sky visibility, we create a state with lidar and
inertial measurements only.

\subsection{Maximum-a-Posteriori Estimation}
We maximize the likelihood  of the measurements $\calZ_k$, given the history of
states $\calX_k$
\begin{equation}
 \X^*_k = \argmax_{\X_k} p(\X_k|\Z_k) \propto p(\X_0)p(\Z_k|\X_k),
\label{eq:posterior}
\end{equation}
where measurements are assumed to be conditionally independent and corrupted by
white Gaussian noise. Therefore, we can express \Equation \eqref{eq:posterior}
as a least-squares minimization~\cite{Dellaert2017}
\begin{equation}
\begin{split}
\X^{*}_k = \argmin_{\X_k}& \|\mathbf{r}_0\|^2_{\Sigma_0} \\
+\sum_{i,j \in \calT_k}  \bigg(&
 \|\mathbf{r}_{\calI_{ij}}\|^2_{\Sigma_{\calI_{ij}}}
 + \|\mathbf{r}_{\calL_{ij}}\|^2_{\Sigma_{\calL_{ij}}} \\
 + \sum_{\rho_i^m\in\calP_i} &\|r_{\rho_i^m}\|^2_{\sigma_{\rho_i^m}} 
 + \sum_{\Delta\Delta\phi^{mn}_{ij}\in\calC_{ij}}
\|r_{\Delta\Delta\phi^{mn}_{ij}}\|^2_{\sigma_{\Delta\Delta\phi^{mn}_{ij}}}
\bigg),
\end{split}
\label{eq:cost-function}
\end{equation}
where $\calT_k$ is the set of all state indices in the sliding smoothing window
up to time~$t_k$. Each term is the residual associated with a factor type,
weighted by the inverse of its covariance matrix. Residuals include a prior, IMU
factors, relative odometry factors from lidar, and two types of GNSS factors,
which are detailed in the following section.

\section{FACTOR GRAPH FORMULATION}

\begin{figure}
\centering
\begin{tikzpicture}[
statenode/.style={circle, draw=black, fill=black!5, very thick, minimum
size=7mm, node distance=8mm},
priornode/.style={circle, draw=black, fill=black, very thick, minimum size=1mm},
priorline/.style={-, draw=black, very thick},
imunode/.style={circle, draw=orange, fill=orange, very thick, minimum size=2mm,
node distance=8mm},
imuline/.style={-, draw=orange, very thick},
landmarknode/.style={circle, draw=green!60, fill=green!5, very thick, minimum
size=7mm},
cpnode/.style={circle, draw=brown, fill=brown, very thick, minimum size=2mm,
node distance=10mm},
cpline/.style={-, draw=brown, very thick},
prnode/.style={circle, draw=purple, fill=purple, very thick, minimum size=2mm,
node distance=10mm},
prline/.style={-, draw=purple, very thick},
lidarnode/.style={circle, draw=blue, fill=blue, very thick, minimum size=2mm,
node distance=10mm},
lidarline/.style={-, draw=blue, very thick},
]
\node[priornode, label={[font=\footnotesize,text=black]below:prior}] (prior) {};
\node[statenode, right of=prior] (x0) {$\State_0$};
\draw [priorline] (prior) -- (x0);
\node[imunode, right of=x0, node distance=16mm] (imu01) {};
\draw [imuline] (x0) -- (imu01);
\node[prnode, above of=x0, xshift=-5mm,
label={[font=\footnotesize,text=purple]above:pseudorange}] (pr00) {};
\draw [prline] (x0) -- (pr00);
\node[prnode, above of=x0, xshift=5mm] (pr01) {};
\draw [prline] (x0) -- (pr01);
\node[cpnode, above of=imu01, xshift=0mm,
label={[font=\footnotesize,text=brown]below:carrier}] (cp010) {};
\draw (x0) edge[cpline] (cp010);
\node[statenode, right of=imu01, node distance=16mm] (x2) {$\State_1$};
\draw [imuline] (imu01) -- (x2);
\draw (cp010) edge[cpline] (x2);
\node[imunode, right of=x2, node distance=16mm,
label={[font=\footnotesize,text=orange]above:IMU}] (imu12) {};
\draw [imuline] (x2) -- (imu12);
\node[prnode, above of=x2, xshift=-5mm] (pr10) {};
\draw [prline] (x2) -- (pr10);
\node[prnode, above of=x2, xshift=5mm] (pr11) {};
\draw [prline] (x2) -- (pr11);
\node[cpnode, above of=imu12, xshift=-5mm] (cp120) {};
\draw (x2) edge[cpline] (cp120);
\node[cpnode, above of=imu12, xshift=5mm] (cp121) {};
\draw (x2) edge[cpline] (cp121);
\node[statenode, right of=imu12, node distance=16mm] (x4) {$\State_2$};
\draw [imuline] (imu12) -- (x4);
\draw (cp120) edge[cpline] (x4);
\draw (cp121) edge[cpline] (x4);
\node[prnode, above of=x4, xshift=0mm] (pr20) {};
\draw [prline] (x4) -- (pr20);
\node[statenode, above of=x2, node distance=20mm] (TEW) {$\T_{\ecef \World}$};
\draw (pr00) edge[prline] (TEW);
\draw (pr01) edge[prline] (TEW);
\draw (pr10) edge[prline] (TEW);
\draw (pr11) edge[prline] (TEW);
\draw (pr20) edge[prline] (TEW);
\draw (cp010) edge[cpline] (TEW);
\draw (cp120) edge[cpline] (TEW);
\draw (cp121) edge[cpline] (TEW);
\node[lidarnode, below of=x2, label={[font=\footnotesize,text=blue]below:lidar %
odometry (ICP)}] (l01) {};
\draw [lidarline] (x0) -- (l01);
\draw [lidarline] (l01) -- (x4);
\end{tikzpicture}
\caption{Factor graph structure with variable nodes (large circles) for states
$\State_i$ and transformation $\T_{\ecef\World}$ between local frame and global
Earth frame and factor nodes (small, colored) for priors, and properioceptive
(IMU), exteroceptive (lidar), and GNSS measurements (pseudorange, carrier
phase).}
\label{fig:factor-graph}
\end{figure}
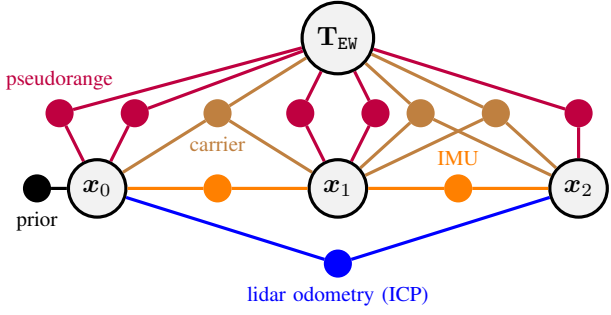
\Figure\ref{fig:factor-graph} shows the factor graph structure. Each measurement
factor is associated with a residual, which is the difference between a
model-based prediction given the connected variables and the observation. We
summarize the IMU and lidar residuals in Sec.~\ref{sec:imu-factors} and
\ref{sec:lidar-factors} before detailing the pseudorange and carrier-phase
residuals in Sec.~\ref{sec:pr} and \ref{sec:cp}.

\subsection{Pre-integrated Inertial Measurements}
\label{sec:imu-factors}
We follow the standard method of IMU measurement pre-integration to constrain
the pose, velocity, and biases between two consecutive nodes $x_i$ and $x_{i+1}=x_j$,
providing high-frequency state updates between nodes.
For a description of the residual $\mathbf{r}_{\calI_{ij}} \in \Real^{15}$, see Forster et al.~\cite{Forster2017a}.

\subsection{ICP Registration}
\label{sec:lidar-factors}

We register the lidar measurements to a local submap~\cite{wisth2022tro} using
iterative closest point (ICP) odometry~\cite{pomerleau2013comparing} and add the
registration output as relative pose factors between times $t_i$ and $t_j$ with
the residual
\begin{equation}
\residual_{\calL_{ij}} = \Phi\left(\tilde{\mathbf{T}}_i^{-1}\tilde{\mathbf{T}}_j^{-1}\mathbf{T}_i^{-1}\mathbf{T}_j\right),
\label{eq:icp-residual}
\end{equation}
where $\mathbf{T}_i=\left[\mathbf{R}_i,\mathbf{p}_i\right]$ is the estimated
pose, $\tilde{\mathbf{T}}_i\in \SEthree$ is the estimate from the ICP module,
and $\Phi$ is the lifting operator defined by Forster et
al.~\cite{Forster2017a}.

\subsection{Pseudoranges}
\label{sec:pr}

For each acquired satellite signal $m$ at time $t_i$, a GNSS receiver reports a
pseudorange $\rho^m_i\in \calP_i$, which is the observed signal travel time
multiplied by the speed of light. It is close to the spatial satellite-receiver
distance, but affected by additional terms, in particular signal delays in
different layers of the atmosphere and time offsets of the clocks that are used
to measure the signal travel time~\cite{kaplan2005understanding}. The
pseudorange residual for a single received satellite signal $m$ is approximately
\begin{equation}
\begin{split}
r_{\rho^m_i} =~&
\|\sat^m_i-\T_{\ecef \World}\mathbf{p}_i\| \\
&+ \delta\rho_\mathrm{T}\left(\sat^m_i,
\T_{\ecef \World}\mathbf{p}_i\right)
+ %
\delta\rho_\mathrm{I}\left(\mathbf{s}^m_i,
\T_{\ecef \World}\mathbf{p}_i\right) \\
&+ \lightspeed \cdot \delta{}t^\mathrm{r}_{i,g}
- \left(\rho^m_i + \lightspeed \cdot (\delta{}t^m + \nu^m)\right),
\end{split}
\label{eq:pr-residual}
\end{equation}
where $\mathbf{s}^m_i \in \Real^3$ is the satellite position in frame~$\ecef$ at
the signal transmit time that corresponds to the observation time $t_i$. It is
corrected for the Earth rotation. The spatial signal delay in the troposphere is
$\delta\rho_\mathrm{T} \colon{} \Real^3 \times \Real^3 \rightarrow{} \Real^+_0$,
the delay in the ionosphere is $\delta\rho_\mathrm{I} \colon{} \Real^3 \times
\Real^3 \rightarrow{} \Real^+_0$, the speed of light is $\lightspeed$, and the
offset of the receiver clock from the clock of the GNSS with index
$g\in\{0\ldots3\}$ is $\delta{}t^\mathrm{r}_{i,g} \in \Real$. The pseudorange
$\rho^m_i$ is adjusted to correct for satellite clock bias $\delta{}t^m\in\Real$
and relativity $\nu^m\in\Real$.

We model satellite position, satellite clock offset, and both atmospheric delays
based on data broadcasted by the satellites. Due to these effects, modeling
errors can be up to a few meters. The antenna position and the receiver clock
offset are the unknowns that we need to estimate.
If we use a multi-band receiver, then there can be multiple residuals for a
single satellite, one for each band in which the receiver acquired a signal.
Alternatively, observations from different bands can be combined to estimate atmospheric delays.

\subsection{Carrier Phases}
\label{sec:cp}

The residual of carrier phase $\phi^m_i\in\Real^+$ (in units of lengths) could
be written similarly to the pseudorange residual~\cite{kaplan2005understanding}
\begin{equation}
\begin{split}
r_{\phi^m_i}&=~
\|\mathbf{s}^m_i-\T_{\ecef \World}\mathbf{p}_i\| \\
&+ \delta\rho_\mathrm{T}\left(\mathbf{s}^m_i,
\T_{\ecef \World}\mathbf{p}_i\right)
- %
\delta\rho_\mathrm{I}\left(\mathbf{s}^m_i,
\T_{\ecef \World}\mathbf{p}_i\right)
+ \lambda^m \omega^m \\
&+ \lightspeed \cdot \delta{}t^\mathrm{r}_{i,g}
- \left(\phi^m_i + \lambda^m N^m_i + \lightspeed \cdot (\delta{}t^m +
\nu^m)\right).
\end{split}
\label{eq:cp-single-residual}
\end{equation}
Most of the terms are the same as in Eq.~\eqref{eq:pr-residual}. However, the
delay in the ionosphere $\delta\rho_\mathrm{I}$ has the opposite sign. There is
also a wind-up effect $\omega^m \in \Real$ resulting from the interplay between
the changing satellite orientation and the circularly polarized carrier wave
(with wavelength $\lambda^m \in \Real^+$), which has a magnitude ranging from
centimeters to a few decimeters~\cite{wu1992effects}. The most significant
difference is an unknown offset in the number $N^m_i\in\mathbb{N}$ of
wavelengths.

The advantage of carrier-phase observations over pseudorange ones is that their
measurement noise is only $\sim$\SI{5}{\milli\meter}. However, modeling them
precisely in real time is challenging because satellite positions and
atmospheric delays are not necessarily known with submeter accuracy.
Furthermore, the integer ambiguity $N^m_i$ is an additional unknown that needs
to be estimated. Therefore, we apply double-differencing, i.e., combine multiple
observations to cancel out terms that are unknown or imprecisely known.
Specifically, we combine two observations of two different satellites of the
same GNSS and signal band at the current time $t_i$ with two older observations
of the same satellites at a time $t_i<t_j$~\cite{suzuki2020time}
\begin{equation}
\begin{split}
\Delta\Delta\phi^{mn}_{ij} =~&
\left(\tilde{\phi}^n_j - \tilde{\phi}^m_j\right) -
\left(\tilde{\phi}^n_i - \tilde{\phi}^m_i\right),
\end{split}
\end{equation}
where $\tilde{\phi}^m_i = \phi^m_i + c \cdot (\delta{}t^m + \nu^m)$ is the
carrier phase $\phi^m_i$ corrected for satellite clock bias and relativity. If
the change of the atmospheric delays of the satellite signals $m$ and $n$
between times $t_i$ and $t_j$ is negligible and the signals are still locked
(i.e., $N^m_i=N^m_j$ and $N^n_i=N^n_j$), then the residual is
\begin{equation}
\begin{split}
r_{\Delta\Delta\phi^{mn}_{ij}} =~&
\left(\|\mathbf{s}^n_j-\T_{\ecef \World}\mathbf{p}_j\| -
\|\mathbf{s}^m_j-\T_{\ecef \World}\mathbf{p}_j\|\right) \\
&- \left(\|\mathbf{s}^n_i-\T_{\ecef \World}\mathbf{p}_i\| -
\|\mathbf{s}^m_i-\T_{\ecef \World}\mathbf{p}_i\|\right) \\
&- \Delta\Delta\phi^{mn}_{ij}.
\end{split}
\label{eq:cp-residual}
\end{equation} %
For each pair of band and satellite system (GPS, GLONASS, Galileo, BeiDou), we
select the satellite that has been continuously visible for the longest time for
the first satellite signal $m$. For the second satellite signal $n$, we iterate
over all remaining satellite observations in the same signal band. We create a
residual for each satellite pair obtained this way.

\begin{table*}
\caption{Mean (top) and median (bottom) horizontal localization errors [m] in
the global Earth frame with RTK as ground truth.}
\centering
 \resizebox{\textwidth}{!}{%
  \begin{tabular}{llr|rrrr|rr}
 \multicolumn{3}{l}{} & \multicolumn{4}{c}{baseline methods} &
\multicolumn{2}{c}{our proposed method} \\
\textbf{dataset} & \textbf{platform}
& \textbf{duration [s]} & \textbf{GNSS-fix} & \textbf{IMU, GNSS-fix} &
\textbf{IMU, ICP, GNSS-fix} & \textbf{GVINS} & \textbf{IMU, raw-GNSS} &
\textbf{IMU, ICP, raw-GNSS} \\
 \hline
  \textbf{HK} & car & 2454 & & & & 1.75 & \textbf{1.54} & \\
  & & & & & & 1.63 & \textbf{1.44} & \\
  \rowcolor{gray!25}
\textbf{Jericho} & car & 923 & 4.01 & 4.04 & failure & & 2.22 & \textbf{2.03} \\
  \rowcolor{gray!25}
  & & & 2.67 & 2.82 & & & 1.88 & \textbf{1.71} \\
  \textbf{Park Town} & car & 264 & 4.57 & 4.46 & & & \textbf{2.79} & \\ %
  & & & 2.56 & 2.54 & & & \textbf{2.00} & \\ %
  \rowcolor{gray!25}
\textbf{Bagley} & quadruped & 1120 & 3.40 & 3.46 & 4.78 & & 2.34 & \textbf{2.07}
\\
  \rowcolor{gray!25}
  & & & 2.84 & 3.16 & 5.50 & & \textbf{1.97} & 2.02 \\
\textbf{Thom} & quadruped & 440 & 12.31 & 10.93 & 4.24 & & 1.66 & \textbf{1.33}
\\
  & & & 7.42 & 9.32 & 2.92 & & \textbf{0.86} & 1.00 \\
 \end{tabular}
 }%
\label{tab:results}
\end{table*}

\section{IMPLEMENTATION}

We integrated the GNSS factors from Sec.~\ref{sec:pr} and~\ref{sec:cp} into the
VILENS state estimator, which includes IMU and lidar registration factors
already and runs in real time~\cite{wisth2022tro}. We created lidar registration factors at half the
rate of the GNSS factors and solved the factor graph with a fixed lag smoother
based on the incremental optimizer iSAM2~\cite{kaess2012isam2} in the GTSAM
library~\cite{Dellaert2017}. The transformation $\mathbf{T}_{\ecef
\World}$ between the local and global frame was continuously estimated, while
the states outside the optimization window $\calT_k$ were marginalized.

Raw GNSS observations are prone to outliers; for example, the multi-path effect
occurs when satellite signals are reflected by surrounding buildings or
vegetation before they are received. This induces a longer travel distance than
the direct line of sight. To mitigate this, we first applied a threshold-based
outlier detector and then used the Huber loss function to reduce the effect of
remaining outliers~\cite{watson2019diss}.

We implemented the GNSS processing components using the GPSTk
library~\cite{harris2007gpstk}. To avoid a cold start of at least
\SI{30}{\second} where satellite positions and clocks are unknown, we preloaded
publicly available satellite navigation data before the start of an experiment.
No online data was used thereafter.

\section{EXPERIMENTAL RESULTS}
\begin{figure}[tb]
    \centering
    \begin{overpic}[height=4.25cm]{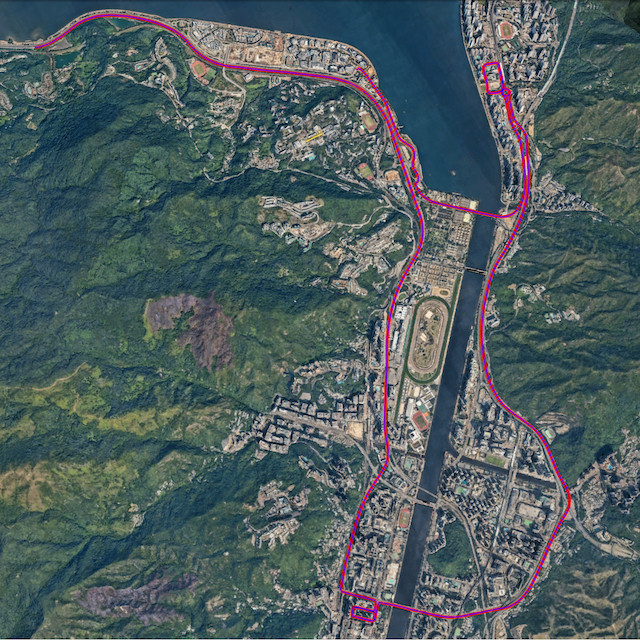}
        \put(2,90){\colorbox{purple!30}{A}}
        \put(2,1){\color{white} \rotatebox{90}{\footnotesize Imagery
\copyright{} Maxar}} \put(49,93){\color{white} \footnotesize start}
            \put(15,89){\color{white} \footnotesize end}
    \end{overpic}
    \begin{overpic}[height=4.25cm]{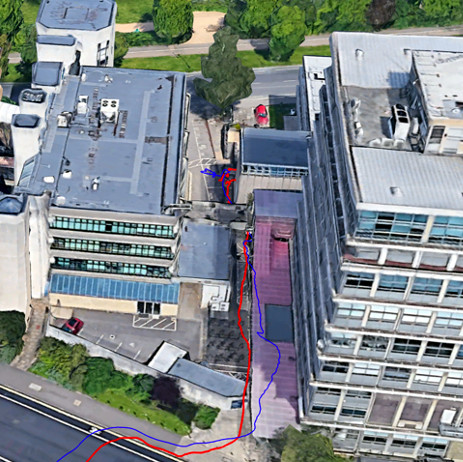}
        \put(0,0){\color{white}
\rotatebox{90}{\colorbox{black!30}{\parbox{0.55in}{\footnotesize Imagery
\copyright{} \\ Landsat \\ Copernicus}}}} \put(2,90){\colorbox{purple!30}{B}}
            \put(30,10){\color{white} \footnotesize start}
            \put(53,66){\color{white} \footnotesize end}
    \end{overpic} \\
    \vspace{1.5mm}
    \begin{overpic}[height=4.25cm]{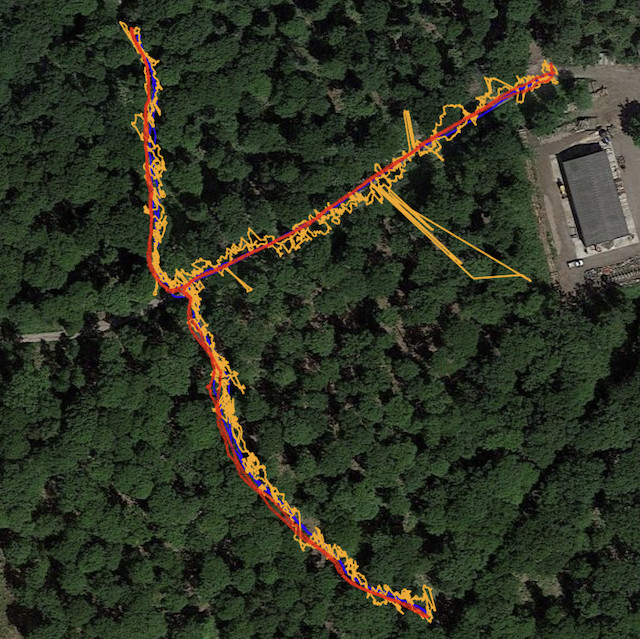}
        \put(2,90){\colorbox{purple!30}{C}}
        \put(2,1){\color{white} \rotatebox{90}{\footnotesize Imagery
\copyright{} Landsat/Copernicus}} \put(71,93){\color{white} \footnotesize
start/end}
    \end{overpic}
    \begin{overpic}[height=4.25cm]{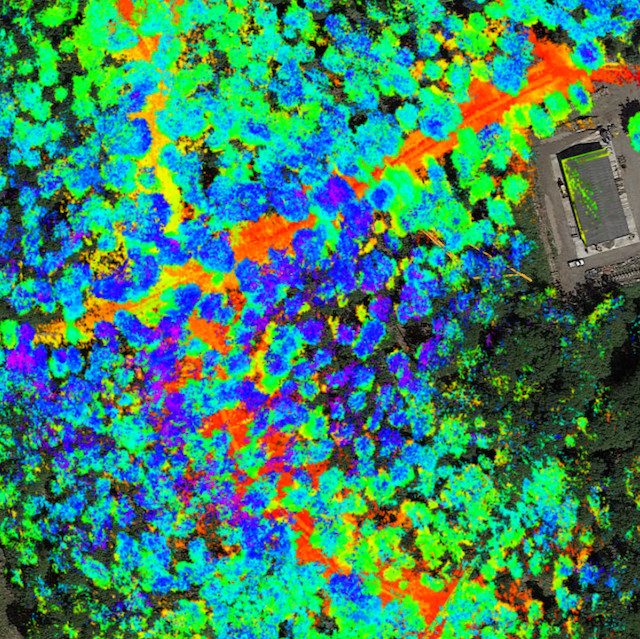}
        \put(2,90){\colorbox{purple!30}{D}}
        \put(2,1){\color{white} \rotatebox{90}{\footnotesize Imagery
\copyright{} Landsat/Copernicus}} \end{overpic} \\
\vspace{1.5mm}
    \begin{overpic}[height=2.95cm]{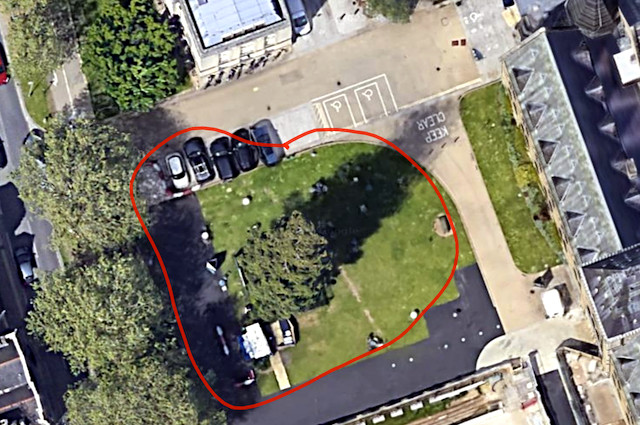}
        \put(2,56){\colorbox{purple!30}{E}}
    \put(2,1){\color{white} \rotatebox{90}{\footnotesize Imagery \copyright{}
Google}} \put(74,50){\color{white} \footnotesize buildings}
            \put(86,48){\color{white} \vector(1,-2){5}}
            \put(72,52){\color{white} \vector(-3,1){30}}
            \put(14,12){\color{white} \footnotesize trees}
            \put(21,18){\color{white} \vector(-1,2){7}}
            \put(28,14){\color{white} \vector(2,1){14}}
            \put(18,52){\color{white} \footnotesize start/end}
            \put(31,51){\color{white} \vector(2,-1){12}}
    \end{overpic}
    \begin{overpic}[height=2.95cm]{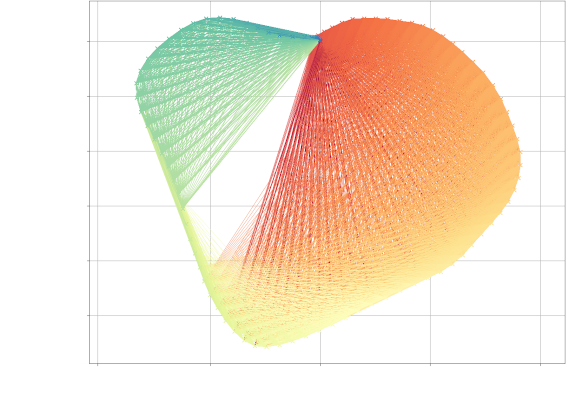}
        \put(2,62.5){\colorbox{purple!30}{F}}
    \put(45,-4){\footnotesize east [m]}
    \put(9,1.5){\footnotesize -20}
    \put(31,1.5){\footnotesize -10}
    \put(55,1.5){\footnotesize 0}
    \put(72,1.5){\footnotesize 10}
    \put(92,1.5){\footnotesize 20}
    \put(0,20){\rotatebox{90}{\footnotesize north [m]}}
    \put(6,15){\footnotesize -20}
    \put(6,25){\footnotesize -10}
    \put(11,35){\footnotesize 0}
    \put(8,45){\footnotesize 10}
    \end{overpic}
\caption{Experimental results. \textbf{(A)}$^1$: Trajectory of a car in Hong
Kong estimated with our algorithm fusing inertial and GNSS measurements (red)
compared to RTK (ground truth, blue)~\cite{cao2022tro}. \textbf{(B)}: Trajectory
of a quadruped traversing between two buildings estimated using IMU, ICP, and
GNSS (red) compared to RTK (blue), part of sequence \textit{Thom}. The RTK
trajectory drifted into the building when there was little GNSS data while our
estimate did not. \textbf{(C)}: Trajectory of a quadruped in the Bagley Wood
estimated using IMU, ICP, and GNSS (red) in comparison to RTK (blue) and single
GNSS fixes (orange). \textbf{(D)}: Same trajectory with lidar scans overlayed.
\textbf{(E)}: Trajectory of a handheld device estimated using time-differential
carrier-phase factors only and given the starting position. \textbf{(F)}: The
corresponding factors between different states in time, represented by lines and
crosses respectively. Few factors were created to states near the trees on the
left. Still, the horizontal localization error at the end was less than 10~cm.}
\label{fig:experiments}
\end{figure}
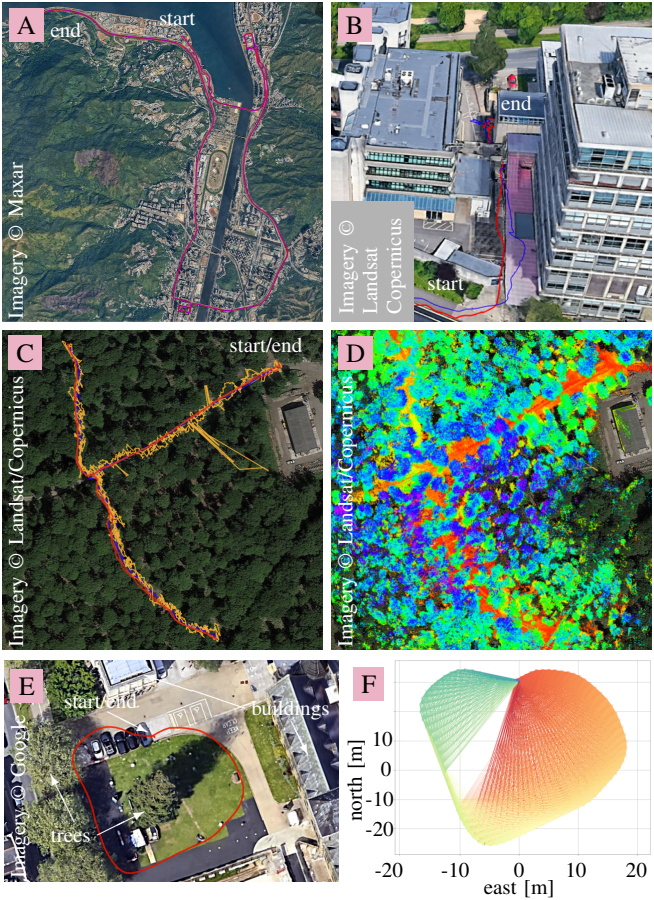

To evaluate our proposed algorithm, we conducted three experiments using both,
public and self-recorded datasets. 

\textit{Experiment~1 (Sec.~\ref{sec:exp1} - raw GNSS and IMU fusion)}: We
compared against the state-of-the-art using the \textit{urban driving} sequence
of the public GVINS dataset~\cite{cao2022tro}. The setup for this dataset
included a \SI{200}{\hertz} consumer-grade \textsc{Analog Devices ADIS16448}
IMU, two \SI{20}{\hertz} \textsc{Aptina MT9V034} cameras, and a \SI{10}{\hertz}
entry-level multi-band \textsc{u-blox C099-F9P} GNSS receiver. Because there was
no lidar data, we evaluated only the fusion of IMU and raw GNSS with our
algorithm. The sequence has a duration of \SI{41}{\minute} and includes sections
with little sky visibility (urban canyons) and brief complete GNSS drop-outs
(underpasses). We also collected four sequences with limited sky visibility of
our own by mounting a \SI{5}{\hertz} \textsc{C099-F9P} GNSS receiver with a
multi-band GNSS antenna and a consumer-grade \SI{200}{\hertz} \textsc{Bosch
BMI085} IMU on an electric vehicle and a \textsc{Boston Dynamics Spot} quadruped
robot, see Fig.~\ref{fig:fg_art}. The sequences evaluated were:
\begin{itemize}
\item \textit{Jericho}: a \SI{4.4}{\kilo\meter} driving loop through narrow
streets in Oxford's city center at speeds up to \SI{11}{\meter\per\second}. This
includes several sections with GNSS drop-out. \item \textit{Park Town}: a
\SI{1.2}{\kilo\meter} drive along tree-lined avenues.
\item \textit{Bagley}: the quadruped moving through a dense commercial
forest. This is known to be challenging for satellite navigation due to the very
limited sky visibility, many outliers in the GNSS measurements because of signal
reflections by surrounding vegetation (multi-path effect), and signal
degradation caused by the electromagnetic interference of the robot with the
GNSS signals. This is quantified by two measures: first, the pseudorange observations of the receiver have on average four-times higher standard deviations and, second, on average 25\% less satellites are visible, both in comparison to \textit{Jericho}.
\item \textit{Thom}: the quadruped walked around a
high-rise building, passed through a tunnel for \SI{45}{\second} 
and ended in a yard between tall buildings, see Fig.~\ref{fig:experiments}-B.
\end{itemize}

\textit{Experiment~2 (Sec.~\ref{sec:exp2} - raw GNSS, IMU, and lidar fusion)}:
We evaluated fusion with data from a \SI{10}{\hertz} \textsc{Hesai XT32} lidar
that was part of the setup for the sequences \textit{Jericho}, \textit{Bagley},
and \textit{Thom}.

\textit{Experiment~3 (Sec.~\ref{sec:exp3} - carrier-phase-only fusion)}:
Finally, we tested the carrier-phase factor's usefulness for accurate local
navigation. For this, we hand carried the GNSS receiver around a \SI{12}{\meter}
tall tree for \SI{105}{\meter} and \SI{2}{\minute},
cf.~\Figure\ref{fig:experiments}-E, and created the sequence \textit{Tree}.

For all sequences, we used RTK to obtain ground truth.

\subsection{Experiment 1: Raw GNSS and IMU fusion}
\label{sec:exp1}

The column \textbf{GVINS} of Tab.~\ref{tab:results} shows the localization
errors of the open-source GVINS algorithm, which fuses pseudoranges and Doppler
shifts from the GNSS receiver with inertial measurements and vision constraints
from a camera. The column \textbf{IMU, raw-GNSS} presents results for our
algorithm fusing pseudoranges, carrier phases, and inertial measurements only.
Our algorithm performs slightly better than GVINS despite the fact that we do
not use a camera. This demonstrates that carrier-phase observations can replace
exteroception for accurate local navigation when the latter is unavailable, but
some sky visibility exists.

There is a practical difference between the algorithms, too: GVINS requires a
dedicated initialization phase with sufficient sky visibility that lasts for
\SI{15.3}{\second} for the \textit{urban driving} sequence. In contrast, our
algorithm does not require such a separate phase and converges to the correct
pose of~$\Base$ in the Earth frame~$\ecef$ as soon as the first slight motion
occurs, after less than \SI{4}{\second}. We assume that this is partially due to
the utilization of carrier phases: a small motion is sufficient to estimate the
orientation of~$\Base$ in~$\ecef$ from the precise differential carrier phases,
while more motion is required to achieve the same with the less accurate
pseudoranges and Doppler shifts.

In Tab.~\ref{tab:results}, we compare the accuracy of separately computed
non-differential GNSS fixes (column \textbf{GNSS-fix}), fusion of these fixes
with inertial measurements (\textbf{IMU, GNSS-fix}), and our own algorithm
fusing raw GNSS observations with inertial measurements (\textbf{IMU, raw-GNSS})
for our sequences. The median global accuracy is \SIrange{1}{2}{\meter}, which
is sufficient for many autonomous vehicle applications, e.g., to initialize a
fine-grained lidar localization system  or to reject incorrect place proposals
from such a system. The accuracy is also close to that achieved on the
\textit{Hong Kong} sequence despite the GNSS receiver operating at half the rate
and our sequences being shorter and having no sections with as good sky
visibility. Both issues hinder global convergence. Furthermore, the trajectory
estimates are smooth after initial convergence, as can be seen in the
supplementary material$^1$. In particular, the \SIrange{30}{60}{\percent}
smaller error on \textit{Bagley}$^1$ demonstrates the advantage of using
individual GNSS observations and inertial measurements in a single optimization
framework as opposed to separating the computation of GNSS fixes and the fusion.
The gains mainly come from improved robustness in this scenario, which has fewer
and more noisy GNSS observations.

To summarize, these results show that the pseudorange factors enable robust
localization in the global Earth frame. In addition, the carrier-phase factors
help to create a locally smooth and accurate trajectory when exteroception is
absent.

\subsection{Experiment 2: Raw GNSS, IMU, and Lidar Fusion}
\label{sec:exp2}
For the sequences \textit{Jericho}, \textit{Bagley}$^1$, and \textit{Thom}, we
also compare the accuracy of fusing separately computed GNSS fixes with IMU
measurements and ICP (\textbf{IMU, ICP, GNSS-fix}) versus our own algorithm when
fusing raw GNSS observations with inertial measurements and ICP (\textbf{IMU,
ICP, raw-GNSS}) in Tab.~\ref{tab:results}. The global accuracy is similar to
\textit{Exp.~1} because lidar measurements only provide local information.
However, we also obtain a georeferenced map of the local
environment, see Fig.~\ref{fig:experiments}-D. Furthermore, the results for
\textit{Thom} show that our algorithm, with its single optimization stage, can
implicitly handle GNSS drop-out and smoothly switch between GNSS-aided and
non-GNSS navigation, cf. Fig.~\ref{fig:experiments}-B.
The baseline methods fusing GNSS fixes drift more in this case, cf. Tab.~\ref{tab:results}.

\subsection{Experiment 3: Carrier-Phase-Only Fusion}
\label{sec:exp3}
Lastly, we tested our algorithm\footnote{We provide open-source code to
reproduce the results in Fig.~\ref{fig:experiments}-E and
\ref{fig:experiments}-F on \url{https://github.com/JonasBchrt/raw-gnss-fusion}
together with the dataset sequences \textit{Bagley} and \textit{Tree} and
interactive maps with the estimated trajectories corresponding to some of the
results in Tab.~\ref{tab:results}.} with only double-differential carrier-phase
factors and without proprioception or exterioception on the \textit{Tree}
sequence, see Fig.~\ref{fig:experiments}-E and \ref{fig:experiments}-F. Here the
double-differenced carrier-phase measurements provided an input roughly
equivalent to relative local odometry. The horizontal positioning error in the
end was less than \SI{10}{\centi\meter}, indicating performance close to
Suzuki's method~\cite{suzuki2020time}, despite that we only used one
type of GNSS factor and had no clear sky visibility, unlike Suzuki in his experiments~\cite{suzuki2020time}.

\section{CONCLUSIONS}
We have presented a novel system to localize a mobile robot in an unknown
environment in real time by tightly fusing GNSS observations, proprioceptive measurements,
and optionally exteroceptive measurements into a single factor graph
optimization. The approach employs raw GNSS observations from individual
satellites instead of pre-computed position fixes to maximize the information
taken from the GNSS receiver. This includes not only pseudorange observations
for absolute positioning on Earth, but also accurate carrier-phase observations
for precise local localization to support the situation where exteroceptive
measurements are either unavailable or degraded. We showed that the proposed
approach improves accuracy by up to several meters in comparison to the
two-stage algorithms where a pre-computed fix is used in the factor graph
instead of raw data---especially if the view of the sky is limited. Typical
median localization errors in the Earth frame are still just
\SIrange{1}{2}{\meter}.

\bibliographystyle{IEEEtran}
\bibliography{IEEEabrv, root.bib}
\end{document}